\documentclass[11pt,letterpaper]{article}
\usepackage[utf8]{inputenc}
\usepackage[T1]{fontenc}
\usepackage{longtable}
\usepackage{float}
\usepackage{wrapfig}
\usepackage{soul}
\usepackage{textcomp}
\usepackage{marvosym}
\usepackage{wasysym}
\usepackage{latexsym}
\usepackage{amssymb}
\usepackage{hyperref}
\usepackage{cogsys}
\usepackage{times}
\usepackage{inconsolata}
\usepackage{color}
\usepackage{epsf}
\usepackage{graphicx}
\usepackage{subfigure}
\usepackage{longtable}

\usepackage{cogsysapa}

\usepackage{enumitem}
\setitemize[0]{leftmargin=0.20in,itemsep=-0.7pt}
\setenumerate[0]{itemsep=-0.5pt,leftmargin=*}
\hypersetup{colorlinks=true, linkcolor=black, citecolor=black, urlcolor=blue}

\cogsysheading{}{2012}{}{09/2012}{12/2012}

\ShortHeadings{Interactive Acquisition of Grounded Word Representations}
              {S.\ Mohan, A.\ H.\ Mininger, J.\ R.\ Kirk, and J.\ E.\ Laird}

\newenvironment{packeditemize}{
\begin{itemize}
  \setlength{\itemsep}{1pt}
  \setlength{\parskip}{0pt}
  \setlength{\parsep}{0pt}
}{\end{itemize}}

\newenvironment{packedenumerate}{
\begin{enumerate}
  \setlength{\itemsep}{1pt}
  \setlength{\parskip}{0pt}
  \setlength{\parsep}{0pt}
}{\end{enumerate}}

\begin{document}
\title{Acquiring Grounded Representations of Words \\ with Situated Interactive Instruction}
\author{Shiwali Mohan}{shiwali@umich.edu}
\author{Aaron H. Mininger}{mininger@umich.edu}
\author{James R. Kirk}{jrkirk@umich.edu}
\author{John E. Laird}{laird@umich.edu}
\address{Computer Science and Engineering, University of Michigan, Ann Arbor, MI USA}
\vskip 0.2in
\begin{abstract}We present an approach for acquiring grounded representations of words
from mixed-initiative, situated interactions with a human
instructor. The work focuses on the acquisition of diverse types of
knowledge including perceptual, semantic, and procedural knowledge
along with learning grounded meanings. Interactive learning allows the
agent to control its learning by requesting instructions about unknown
concepts, making learning efficient. Our approach has been
instantiated in Soar and has been evaluated on a table-top robotic arm
capable of manipulating small objects. 
\end{abstract}

\section{Introduction}
\label{sec-1}
A goal of our research is to develop autonomous agents that can
dynamically extend not only their knowledge of the world and their
ability to interact with it, but also their use of language for
interacting with humans. We describe initial progress in developing
such an agent that learns grounded representations of adjectives,
nouns, prepositions, and verbs through \emph{interactive situated
  instruction} with a human mentor while performing a task. The agent
is developed in Soar \cite{Laird2012} using the same architectural
mechanisms used in other cognitive tasks without modification. 

Our first claim is that in this approach, learning can exploit a broad
range of information: knowledge extracted from perception; learned
semantic, procedural, and episodic knowledge; and action-model
knowledge. Our second claim is that the learning is incremental
(learned words aid in learning additional words), online (learning
occurs during performance without offline processing), and fast (very
few examples are required to teach new words). The agent is reactive;
the responses to the human instructor are generated in real time (< 2
seconds). These claims are established by demonstration; we create an
agent using this approach that learns via interaction with a human,
determine what sources of knowledge it uses in learning, and evaluate
the characteristics of its learning.

Our remaining claims focus on the properties of situated interactive
instruction. One claim is that situated instructions is an effective
method for learning grounded representations of words that combine the
words with perceived regularities in the environment. The agent has a
vision system for perception and an arm for manipulating objects in
its world. For nouns and adjectives, the agent learns new
classifications of perceptual features (color, size, and shape) from
interactive training with a human and also learns to associate these
classifications with specific words (such as \emph{red}, \emph{large},
or \emph{cylinder}). For prepositions, the agent learns to associate
combinations of primitive spatial predicates (such as the alignment of
objects) with new words (such as \emph{right of}), and for verbs, the
agent learns to associate sequences of primitive actions (such as
\emph{pick up} and \emph{put down} with new words (such as
\emph{move}). The agent begins with no prior knowledge of these nouns,
adjectives, prepositions, and verbs, except for some limited
part-of-speech knowledge used in parsing. As the agent learns, the
concepts are grounded in its experiences and are dependent on the
specifics of the training it receives. For example, the word
\emph{red} could be associated with shape instead of color if that is
what the instructor wishes. This claim is evaluated by teaching and
testing the agent on a variety of nouns/adjectives, prepositions, and
verbs.

Our final claim is that interactive instruction is flexible and
efficient. By using interactive instructions, an instructor can be
freed from using a specific ordering to teach the agent new words and
concepts. Often in human controlled interactive learning such as
learning by demonstration, the onus is on the instructor to provide
good examples from the feature spaces so that the agent can acquire
general hypotheses. The instructor must attempt to build and maintain
an internal model of what the agent knows and doesn't know. This is
especially challenging when the agent is dynamically creating
categories from real-world data. In contrast, with mixed initiative
interaction, the instructor can rely on the agent to initiate an
interaction when needed. This approach can speed instruction by
eliminating the need for the instructor to carefully structure the
interaction or repeatedly check with the agent to ensure it has
completely learned a concept. The agent can actively seek examples of
concepts that are hard to learn and avoid asking for multiple examples
of easily acquired concepts. The instructor can take initiative in
presenting interesting examples to the agent that it might have
overlooked, refining agent's learning. We evaluate this claim by
demonstrating that the agent initiates appropriate interactions to
acquire missing knowledge and comparing the number of examples
collected by the agent for different concepts while learning.

The rest of the paper is organized as follows. In Section \ref{sec-2},
we give an overview of our system including the perceptual and
actuation components of our agent. Section \ref{sec-3} provides a
brief background of the Soar cognitive architecture emphasizing the
mechanisms relevant to our agent design. An overview of the agent
design and different phases in learning with instruction are in
Section \ref{sec-4}. Section \ref{sec-5} describes our human-agent
interaction component that forms the basis of learning with
instruction. In Sections \ref{sec-6}, \ref{sec-7}, and \ref{sec-8}, we
describe acquisition of perceptual nouns and adjectives, spatial
prepositions, and action verbs. We present the empirical evaluation of
the system in Section \ref{sec-9}. We conclude with a discussion on
related work in Section \ref{sec-10}, and future directions in
Section \ref{sec-11}.

\section{System Overview}
\label{sec-2}
\begin{figure}[t]
\begin{center}
\includegraphics[width=0.9\textwidth]{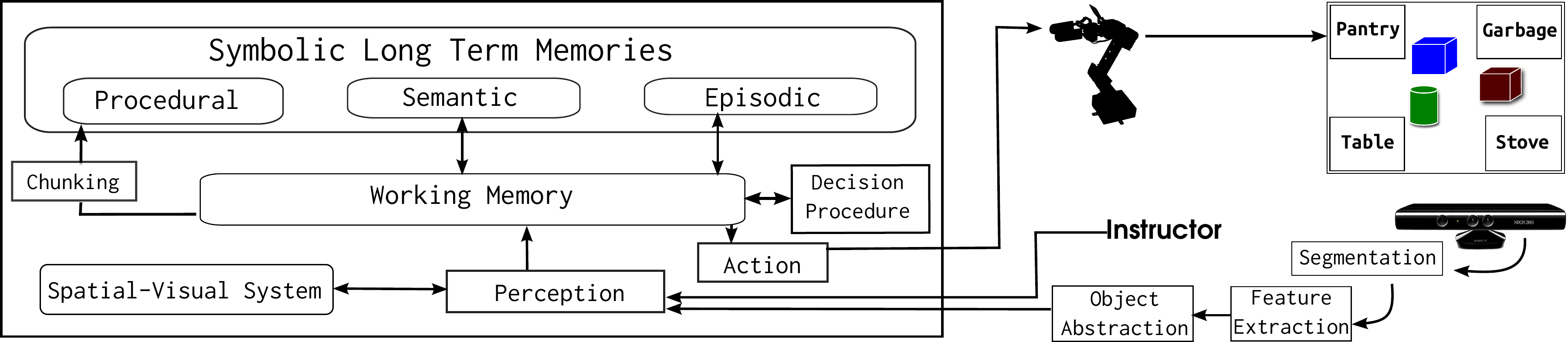}
\caption{System overview including major Soar components.}
\label{fig:soar9}
\end{center}
\vskip -0.3in
\end{figure}

Our agent exists in a simple table-top environment\footnote{The robot
  and the perception/actuation system was developed
  by Edwin Olson at University of Michigan.}
with a robot arm, Kinect sensor, and four predefined locations - a
\emph{stove}, a \emph{dishwasher}, a \emph{garbage} and a
\emph{pantry}. There are a variety of simple foam blocks of different
colors, sizes, and shapes that the arm can manipulate. Figure
\ref{fig:soar9} shows the environment and how the agent interfaces
with it. 

\vspace{-0.15cm}
\begin{packeditemize}
    \item \textbf{Perception}: The perception system segments the
      scene into objects using colored 3D point cloud data provided by
      an overhead Kinect camera. For each of the three perceptual
      properties, features are extracted and classified using a
      K-Nearest Neighbor (KNN) classifier with Gaussian
      weightings. The classification results in a perceptual symbol
      previously learned by the agent. As an example, a perceptual
      symbol \texttt{R43} is associated with the region in the color
      feature space that corresponds to the word \emph{red}. These
      symbols, along with position and bounding box information, are
      used to create a symbolic representation of the object, which is
      provided to the agent. The classifiers are trained exclusively
      through instruction; the process is described in Section
      \ref{sec-6}.

    \item \textbf{Actuation}: To act in the world, the agent sends
      discrete commands to the robot controller. The current commands
      include: \texttt{point-to(object)}, \texttt{pick-up (object)},
      \texttt{put-down (x,y)}. The robot controller then calculates
      and performs the motor actions to execute these  commands. 

    \item \textbf{Instructor Interface}: The instructor interacts with
      the agent through a simple chat interface. Messages to the agent
      are fed through the Link-Grammar parser \cite{Sleator1995} to
      extract part of speech tags and sentence structure. Messages
      from the agent are converted to natural language using
      templates. The instructor can select an object by clicking on it
      in a live camera feed, and the selection is made known to the
      agent. 

\end{packeditemize}

\section{Soar Cognitive Architecture}
\label{sec-3}

Our agent is implemented in Soar, which has been applied to a wide
variety of domains and tasks, including natural language understanding
and robot control. Recent extensions to Soar, including episodic and
semantic memories, as well as a visual-spatial system, enhance Soar's
ability to support grounded language learning. Relevant components are
described in the following paragraphs.

Soar contains a task-independent \textbf{spatial visual system} (SVS)
that supports translations between the continuous representations
required for perception and the symbolic, relational representations
in Soar. The continuous environment state is represented in SVS as a
scene graph composed of discrete objects and their continuous
properties. Binary spatial predicates are computed when an agent
issues a query for a specific predicate such as
\texttt{X-axis-aligned(A,B)}. The set of predicates is task
independent and fixed, but predicate extraction is controlled using
task-specific knowledge. 

\textbf{Working memory} maintains symbolic relational representations
of current and recent sensory data, current goals, and the agent's
interpretation of the current situation including mappings between
objects in the scene and internal symbols and words. Working memory
buffers provide interfaces to Soar's long-term memories, the
perception and action systems, and the instructor interface.

\textbf{Procedural memory} contains Soar's knowledge of how to select
and perform actions (called operators), encoded as if-then rules. The
locus of decision making is not the selection of a rule. Instead, Soar
fires all rules in parallel. The rules propose, evaluate, or apply
operators, which are the locus of decision making. Only a single
operator can be selected at a time, and once an operator is selected,
rules sensitive to its selection and the current context perform its
actions (both internal and external) by modifying working
memory. Whenever procedural knowledge for selecting or applying an
operator is incomplete or in conflict, an impasse occurs and a
substate is created in which more reasoning can occur, including task
decomposition, planning, and search methods. In Soar, complex behavior
arises not from complex, preprogrammed plans or sequential procedural
knowledge, but from the interplay of the agent's knowledge (or lack
thereof) and the dynamics of the environment. In our agent, procedural
memory holds rules that implement the processing capabilities such as
lexical processing, human-agent interaction, grounded comprehension,
and acquisition of grounded representations of words. The agent also
has rules that implement the primitive actions and their models. The
acquired action-execution knowledge for verbs is stored in procedural
memory.

\textbf{Chunking} is a learning mechanism that creates rules from the
reasoning that occurred in a substate. When a result is created in a
substate, a rule is compiled. The conditions of this rule are the
working-memory elements that existed before the substate and were
necessary for creating the result, and the actions are the result. The
rule is added to procedural memory and is immediately
available. Chunking is the mechanism that learns the action-execution
knowledge for novel verbs.

\textbf{Semantic memory} stores context independent declarative facts
about the world. The agent can store working memory elements into
semantic memory and it can retrieve them by creating a cue in a
working memory buffer. The best match to the cue (biased by recency
and frequency) is retrieved from semantic memory to working memory. In
our agent, semantic memory stores \emph{linguistic mapping} knowledge,
such as the mapping between a word and a perceptual symbol (red color
corresponds to symbol \texttt{r43}). Apart from linguistic mapping
knowledge, semantic memory also stores \emph{compositions} of spatial
primitives and \emph{action-concept networks} (discussed later). One
advantage of semantic memory over procedural memory is that any aspect
of a memory can be used for retrieval, whereas in procedural memory,
there is an asymmetry between the conditions and actions. An agent can
use \emph{red} as a cue, or it could use \texttt{r43} as a cue,
depending on what knowledge is available and what knowledge it needed
to retrieve.

\textbf{Episodic memory} stores context-dependent records of the
agent's experiences. It takes snapshots of working memory (episodes)
and stores them in chronological order, enabling the agent to retrieve
both the context and temporal relations of past experiences. The agent
can deliberately retrieve an episode by creating a cue in a working
memory buffer. The best partial match (biased by recency) is retrieved
and added to working memory. Episodic memory facilitates acquisition
of action-execution knowledge through retrospective forward projection
by automatically encoding all interactions accompanied by changes in
sensory perception. The agent can review past instructions and observe
the resulting changes in the environment and its own internal state.

\section{Agent Overview}
\label{sec-4}
\begin{figure}[t]
\begin{center}
\includegraphics[width=1\textwidth]{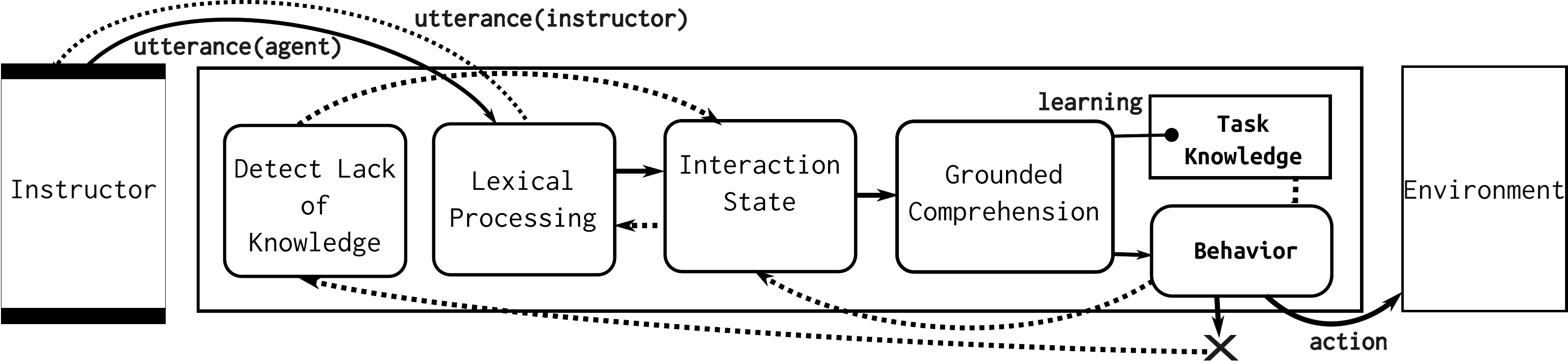}
\vskip -0.1in
\caption{Phases in the interaction cycle.}
\label{fig:interaction-cycle}
\end{center}
\vskip -0.3in
\end{figure}

The agent has procedural knowledge for many types of processing
including perceptual processing, initiating actions, dialog
management, grounded linguistic processing, and learning. During an
\emph{interaction cycle,} the agent selects and applies operators to
interpret instructor's utterances and generate behavior. The
interaction cycle begins with a natural language utterance from the
instructor (shown using solid line in Figure
\ref{fig:interaction-cycle}) and is processed in following phases.

\begin{packedenumerate}
     \item \textbf{Lexical Processing}: LG-Soar \cite{Lonsdale2006},
       a natural language component implemented as operators in Soar,
       generates a syntactic parse of the utterance using a static
       dictionary and grammar. It uses part-of-speech tags to create a
       parse in the agent's working memory, identifying the useful
       content in the message. This parse is further categorized as
       \texttt{verb-command},  \texttt{goal-description},
       \texttt{descriptive-sentence}, etc. based on its lexical
       structure. 

     \item \textbf{Interaction Management}: After the utterance has
       been categorized, it is interpreted within the context of the
       ongoing dialog between the instructor and the agent. Using the
       context and the heuristically determined intentions of the
       utterance, the agent creates a goal to pursue. The goal may
       include performing actions in the environment in response to
       commands from the instructor or providing responses to
       instructor's queries. Interaction management is done using the
       mixed-initiative interaction system described in Section
       \ref{sec-5}. 

     \item \textbf{Grounded Comprehension}: To gain useful information
       from an instruction, the agent must ground linguistic
       references to objects, spatial relationships, and actions. We
       use the term \emph{map} for structures in semantic memory that
       encode how linguistic symbols (nouns/adjectives, spatial
       prepositions, and action verbs) are associated with perceptual
       symbols, spatial compositions, and action-concept
       networks. Maps are learned through interaction with the
       environment and the instructor and are stored in semantic
       memory. To ground a sentence, the \emph{indexing} process
       \cite{Mohan2012c} attempts to retrieve relevant maps from
       semantic memory so that it can connect the linguistic terms
       with their referents. If the terms are successfully mapped, the
       agent uses constraints derived from the retrieved maps, the
       current environment, action models, and the interaction context
       to create a grounded representation of the instruction. These
       sources of knowledge resolve several ambiguities that can arise
       from incomplete referring expressions and polysemous verbs in
       human-agent interactions. If indexing fails
       to retrieve a map or there is insufficient knowledge to resolve
       the ambiguity, an impasse will arise (See Phase 5).  

     \item \textbf{Behavior}: If the agent is successful in generating
       a grounded representation of the instructor's utterance, it
       attempts to pursue the goal associated with the utterance in
       Phase 2. A natural language command \emph{Pick up the red
         triangle} results in the agent picking up the referenced
       object. Apart from actions, the agent can also generate
       descriptions of the scene and can be queried about various
       objects and spatial relationships to verify its learning. If
       there is no action-execution knowledge associated with a
       command, an impasse will arise (See Phase 5). 

     \item \textbf{Impasse and Acquisition}: If the agent fails to
       generate a grounded representation or is unable to execute a
       command, the agent has insufficient knowledge and an impasse
       arises. In response to the impasse, the agent initiates a new
       interaction with the instructor (shown using dotted lines in
       Figure \ref{fig:interaction-cycle}) whose purpose is to acquire
       the missing knowledge. If there are multiple failures during
       interpretation of a new instruction, the agent processes them
       one at a time, leading the instructor through a series of
       interactions until the agent can resolve all the impasses. From
       an impasse arising in Phase 3 and the ensuing interactions, the
       agent can learn maps for nouns/adjectives (Section
       \ref{sec-6}), spatial prepositions (Section \ref{sec-7}), and
       action verbs (Section \ref{sec-8}) leveraging the structure of
       interactions.  Further semantic information about words
       including perceptual symbol categories (Section \ref{sec-6})
       and spatial compositions (Section \ref{sec-7}) are also
       acquired from impasses arising in Phase 3.  Action-concept
       network and execution knowledge (Section \ref{sec-8}) is
       acquired through interactions initiated due to impasses in the
       behavior execution phase.

\end{packedenumerate}

\section{Mixed-Initiative Interaction}
\label{sec-5}
The agent must maintain a state and context of interactions to
comprehend instructions and learn from them, as it is acting in the
environment. Thus, the agent needs a model of task-oriented
interactive instruction. The interaction model must support the
requirements in Table \ref{tbl:requirements}.

\begin{table}[h]
\vskip -0.15in
\caption{Requirements of Mixed Initiative Interaction Model.}
\vskip 0.1in
\label{tbl:requirements}
\begin{small}
\begin{center}
\begin{tabular}{llp{9.6cm}}
\hline
\abovespace\belowspace
\textbf{Requirement}  &  \textbf{Type}            &  \textbf{Description}   \\                                              
\hline\abovespace

 Integrative           &  \textbf{I1}              &  Integrate capabilities, allows agent to plan over and reason about a combined space of linguistic processing, behavior, and learning.                 \\
 Flexible              &  \textbf{F1}: Initiation  &  Both instructor and the agent can direct the interaction.                                                                    \\
                       &  \textbf{F2}: Knowledge   &  Accommodates communication regarding nouns/adjectives, prepositions, verbs, and related questions.                           \\
 Task-oriented         &  \textbf{T1}: Contextual  &  Captures discourse context useful for comprehending incomplete sentences, resolving referent ambiguities, and learning verbs.  \\
                       &  \textbf{T2}: Relevant    &  Instruction-oriented interpretation of human utterances. Agent can ask task-relevant queries.                         \\
\belowspace
                       &  \textbf{T3}: Structural  &  Organizes dialog so that it is useful in task execution and learning.                                               \\
\hline
\end{tabular}
\end{center}
\vskip -0.10in
\end{small}
\end{table}

\begin{figure}[t]
\begin{center}
\includegraphics[width=0.95\textwidth]{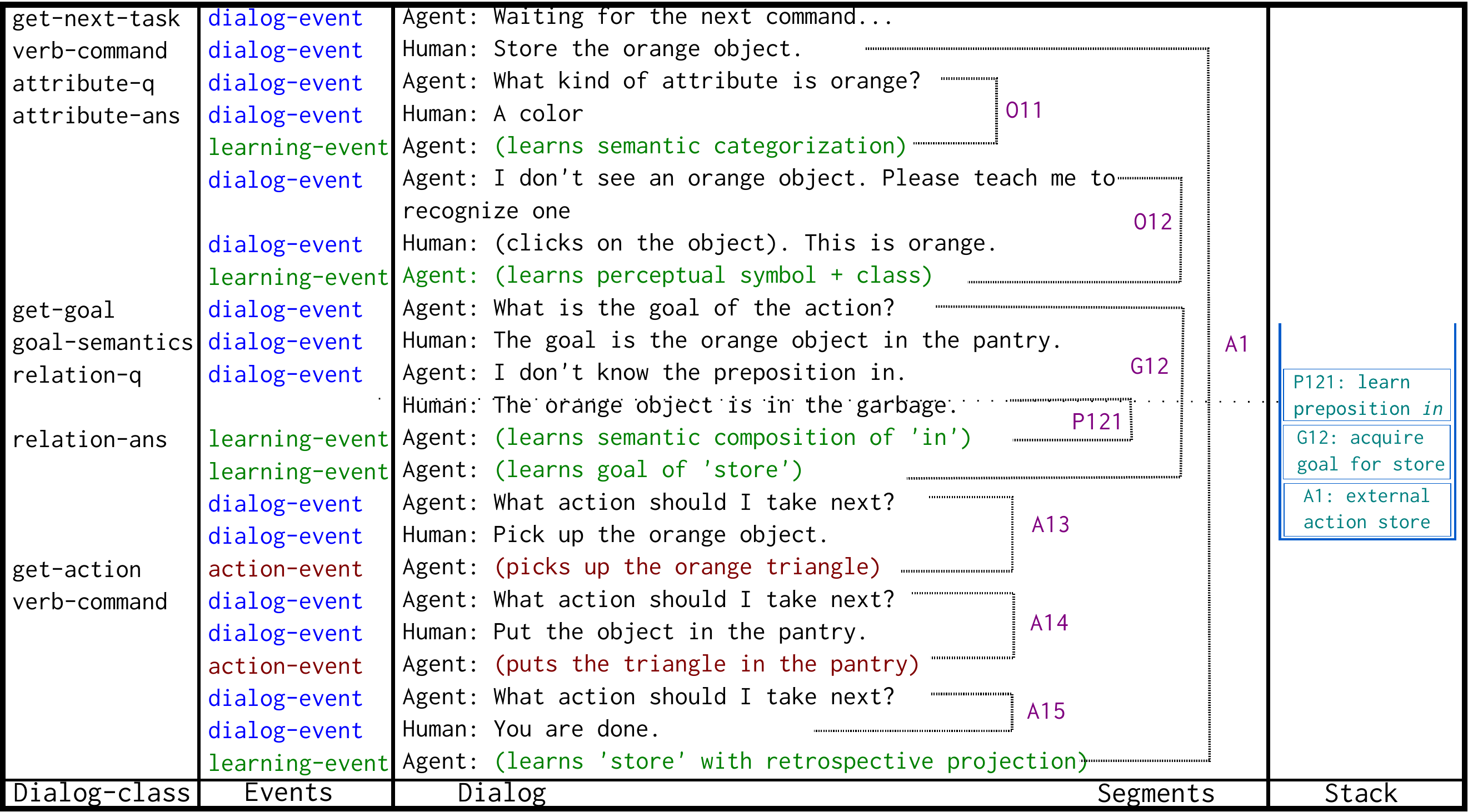}
\caption{Annotated human-agent dialog for acquisition of \emph{store}}.
\label{fig:dialog}
\end{center}
\vskip -0.35in
\end{figure}

Our interaction model is based on a theory of discourse structure that
stresses the role of purpose in discourse and has been adapted from
\emcite{Rich1998a}, extending their framework to accommodate learning
from situated instruction. The state of interaction is represented by
\emph{events,} \emph{segments,} and the \emph{interaction stack.}
Figure \ref{fig:dialog} is an annotated trace of how these concepts
are used by the agent while learning the verb \emph{store.} The agent
also needs to learn the adjective \emph{orange} and preposition
\emph{in} in order to learn the verb.

\vspace{-0.25cm}
\begin{packeditemize}
\item \textbf{Event}: An event causes change, either in the
  environment (\emph{action-event}), the discourse state of the
  instructor-agent interaction (\emph{dialog-event}), or the agent's
  knowledge (\emph{learning-event}). Action-events correspond to
  actions that occur in the environment. Utterances are categorized as
  dialog-events which are assigned to different classes based on their
  lexical and syntactic structures, such as \texttt{get-next-task},
  \texttt{verb-command}, and \texttt{attribute-query} as shown in the
  left most column of Figure \ref{fig:dialog}. A learning-event is the
  successful acquisition of linguistic mapping, semantic, or
  procedural knowledge. The state space of interactions is defined
  over utterances, actions, and learning, addressing the
  \emph{Integrative} (I1) requirement.

\item \textbf{Segment}: A discourse segment is a contiguous sequence
  of events that serves a specific \emph{purpose} and organizes a
  dialog into purpose-oriented interactions. When initiated by the
  agent, the segments allow the agent to learn a new verb
  (\texttt{A1}), acquire a mapping for a novel word (such as
  \texttt{O11} and \texttt{P121}), acquire a goal (\texttt{G12}), or
  acquire an action sequence that is required to execute a verb
  (\texttt{A13}, \texttt{A14}, \texttt{A15}). The segments provide the
  context for the agent to organize its processing and interactions in
  pursuit of its task goals (requirement T3). We have encoded
  heuristics specific to the problem of learning with instruction
  (requirement T2) that are useful in determination of the purpose (a
  set of events) of a segment. These heuristics also influence how the
  agent interprets instructor utterances. Sentences such as \emph{The
    orange object is in the garbage} are treated as learning
  examples. The segments also encode the reason why they were
  initiated (requirement \texttt{T1}) which informs language parsing,
  comprehension, and learning. The context of the segment \texttt{O11}
  is useful when parsing the noun-phrase fragment and learning the
  word \emph{orange}.

\item \textbf{Stack}: The attentional structure of discourse is
  captured in a stack of active segments. When a new segment is
  created, it is pushed onto the stack. The top segment is the focus
  of the current interaction, and the agent acts to achieve its
  purpose. When the purpose of the top segment is achieved, it is
  popped from the stack. The right-most column in Figure
  \ref{fig:dialog} shows a snapshot of the stack.  It contains three
  open segments, \texttt{P121}, \texttt{G12}, and \texttt{A1}, with
  \texttt{P121} being the top segment. The segments are hierarchically
  ordered with each segment contributing towards achieving its
  parent's purpose (which is lower in the stack). To learn
  \emph{store}, the agent must acquire a description of the goal and
  must learn the spatial concept corresponding to the preposition
  \emph{in}. The stack captures the current context of the dialog
  (aiding in requirement \texttt{T1}). The model allows any
  participant to initiate a segment on the stack, supporting the
  \emph{flexibility} requirement, \texttt{F1}.
\end{packeditemize}

\section{Perceptual Noun/Adjective Acquisition}
\label{sec-6}
A key step in the Grounded Comprehension phase (Section 4) is the
resolution of references made to objects in the environment. Our agent
can resolve \emph{gestural} references (selecting an object in the
camera feed), \emph{descriptive} references (noun phrases like
\emph{the large triangle}), and \emph{spatial} references (\emph{the
  triangle in the pantry}). To ground nouns/adjectives used in
descriptive references the agent must connect each word to a set of
features in the perceptual system.

For each perceptual property (e.g., color), different features are
extracted by the perceptual system to create a separate feature
space. For example, the feature space for color is the average RGB
values of the object's point cloud. Through instruction, a classifier
is trained in each feature space and it partitions the space into
regions, each corresponding to a perceptual symbol used by the
agent. The agent must learn both a \emph{linguistic mapping} (the word
to a perceptual symbol) and a \emph{perceptual mapping} (the
perceptual symbol to a region in a feature space).

\subsection{Characterization}
\label{sec-6-1}
Our initial focus is on nouns/adjectives, such as \emph{orange},
\emph{large}, and \emph{triangle}, that describe the perceptual
properties: color, size, and shape. Objects are assumed to be
monochromatic and so can be described by single colors. Sizes are
considered uniform across objects, and relative sizes between objects
are not considered. Shapes are limited to those with a distinctive
outline from above.

The learning of new nouns and adjectives is \emph{fast}; it requires
only a few interactions. This results from several factors, including
the characteristics of the objects used, the classification algorithm,
the interaction model, and the use of engineered feature spaces for
each perceptual property. The use of highly specific feature
extractors means we cannot teach perceptual nouns/adjectives not
easily captured by one of those feature spaces.  Our approach is also
\emph{incremental}. For example, as soon as a color is taught, it can
be used to distinguish objects when teaching new verbs or
prepositions. Splitting the feature space into specific properties
also allows the agent to compose adjectives to recognize objects it
has not seen before. 

\subsection{Background Knowledge}
\label{sec-6-2}
\textbf{Linguistic}: The agent begins with knowledge that
\emph{color}, \emph{shape}, and \emph{size} are perceptual
properties. The agent does not start  any knowledge of specific
nouns/adjectives or perceptual symbols. 

\textbf{Perceptual}: The perceptual system has pre-encoded knowledge
about how to extract useful features for each of the three perceptual
properties. Within each property's feature space, the perceptual
system starts with no initial partitioning and no pre-trained data.

\subsection{Acquisition}
\label{sec-6-3}
\textbf{Linguistic}: When presented with a new noun/adjective like
\emph{orange}, the agent initiates an interaction to learn which of
the three perceptual properties it describes. This knowledge is
acquired through an explicit interaction with the instructor (Figure
\ref{fig:dialog}, \texttt{O11}). The response from the instructor
along with the word is used to create a new perceptual symbol
(eg. create the symbol \emph{c31} for the color orange). A mapping
from the word to the perceptual symbol is stored in semantic
memory. We use perceptual symbols instead of the words directly when
communicating with the perceptual system in order to add a layer of
indirection. This layer gives future flexibility if two words have the
same perceptual grounding (synonyms) or one word has multiple
groundings (orange as a shape or color). 

\textbf{Perceptual}:
When the instructor uses a noun or adjective to describe an object
(e.g. \emph{This is orange}, Figure \ref{fig:dialog}, \texttt{O12}),
the agent uses the linguistic mapping to get a perceptual symbol. It
then tells the visual system to use that object as a training example
for the corresponding perceptual symbol and its property (e.g. use
obj7 as an example of \emph{c14} in the color classifier). This
example refines the classifier and the region in the feature space
corresponding to the perceptual symbol, thus improving the agent's
perceptual knowledge. Our approach does not depend on any particular
type of classifier. A KNN is used because it easily incorporates new
training examples, it handles multiple classes, and it allows a
perceptual symbol to connect to disjoint and non-linear regions of the
feature space.

\section{Spatial Preposition Acquisition}
\label{sec-7}

In order to interact naturally with objects in the world, the agent
needs to understand their relative positions with respect to other
objects and locations. To interact with a human mentor, the agent must
understand how prepositions, such as \emph{next to}, \emph{in}, or
\emph{to the right of}, connect to these spatial
relations. Understanding spatial relations and their connection to
language is also important to executing commands in the world, where
the goal is often to change spatial relations between objects.

The spatial preposition acquisition process supports many modes of 
learning and interaction with prepositions. The human user can teach a
new preposition with a grounded example in the world (\emph{The
  triangle is left of the square.}), query the system about current
relations (\emph{What is left of the square?}), resolve ambiguity
(\emph{[Which block?]-The one in the pantry}), use learned
prepositions in actions (\emph{Put the square in the pantry.}), and
track spatial goals of actions (\emph{The goal is the square in the
  pantry.}) These capabilities require grounding the linguistic
prepositional term to a representation of the appropriate spatial
relation as well the ability to use that information for
comprehension, communication, and action.

The learning of spatial relationships and their associated
prepositions depends on spatial, semantic, and linguistic
knowledge. Spatial knowledge, in the form of primitive spatial
relationships, is built into the system, while the logical combination
of spatial primitives (semantic knowledge) and the mapping of the
prepositional term to this combination (linguistic knowledge) are
learned. Both the semantic and linguistic knowledge are represented as
a network in agent's semantic memory. In the following sections, we
characterize preposition acquisition and give implementation details.

\subsection{Characterization}
\label{sec-7-1}
The training example of a preposition requires three components: a
primary object, a reference object, and a linguistic term for the
preposition. In the teaching statement \emph{the red object is left of
  the blue object}, the \emph{red object} is the primary object, the
\emph{blue object} is the reference object, and \emph{left of} is the
preposition. The agent learns only one perspective, the one the mentor
is using, and cannot change perspective unless the camera is moved or
the prepositions are retaught. The agent can acquire prepositions that
are characterized along the following dimensions.

\vspace{-0.15cm}
\begin{packeditemize}
\item Primitives: The agent can learn prepositions that are
  compositions of two types of primitives.   

\begin{packeditemize}
\vspace{-0.15cm}
\item Direction: The directional primitives describe how the reference
  and primary objects are aligned along each axis in a 3-dimensional
  coordinate system: \texttt{X}, \texttt{Y}, and \texttt{Z}. In
  relation to the reference object, the primary object can be
  \texttt{aligned}, \texttt{greater-than}, or \texttt{less-than}. For
  example two objects that are \texttt{Z-aligned} are on the same
  plane. These relations are useful in learning prepositions that are
  based on spatial order, such as \emph{right of} or \emph{diagonal
    with}.

\item Distance: These primitives encode the distance between the
  reference and the primary object along each axis. The distance is
  measured from the closest surface of each object. Distance-based
  primitives are useful in the acquisition of prepositions such as
  \emph{near} or \emph{far}.

\end{packeditemize}
\vspace{-0.15cm}
\item Composition: The learned spatial relations for prepositions are
  represented by a logical combination of directional primitives and a
  distribution of distance-based primitives.  The combination of
  directional primitives contain \emph{conjunctions} from different
  axes, such as \texttt{X-less-than} \emph{and} \texttt{Z-aligned},
  and \emph{disjunctions} on the same axis, such as \texttt{Y-aligned}
  \emph{or} \texttt{Y-greater-than}. The initial teaching
  demonstration results in a representation with a conjunction of the
  current true directional primitives. Subsequent demonstrations can
  add disjunctive primitives. Additional demonstrations provide a
  distribution of distances from which a range can be
  calculated. Logical combinations of primitives allow the agent to
  acquire a wide range of complex spatial prepositions, including ones
  based on both distance and direction such as \emph{next-to}.

\end{packeditemize}
The acquisition process is both fast and general, learning a large
variety of useful prepositions from a small number of examples. This
learning is also incremental, both in how it uses the knowledge gained
from the other learning mechanism, like nouns and adjectives, and in
how the representation of spatial primitives is refined with
additional training examples. 
\begin{figure}[h]
\begin{center}
\includegraphics[width=0.8\textwidth]{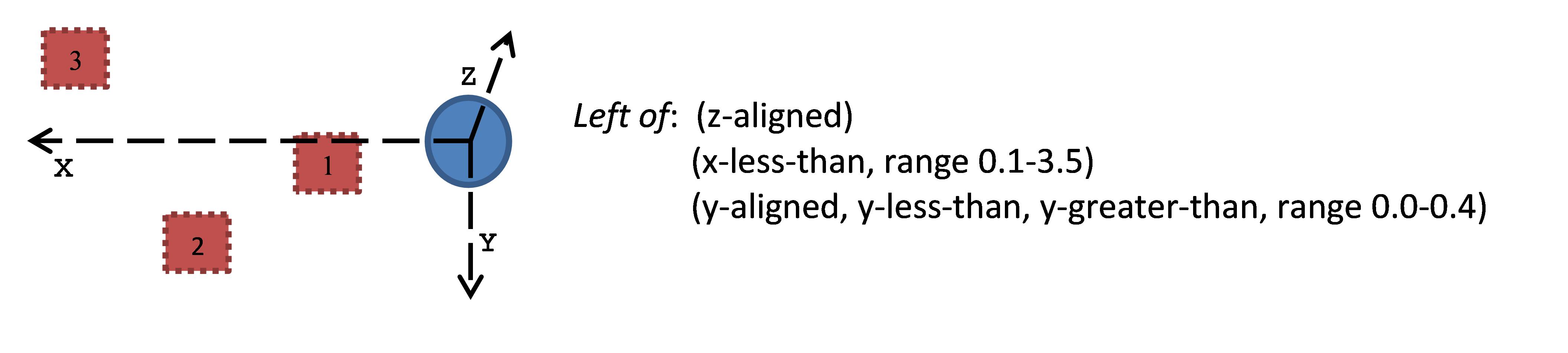}
\vskip -0.2in
\caption{Top down view of objects and the representation of \emph{left of} learned from the three marked examples.}
\label{fig:leftmap}
\end{center}
\vskip -0.3in
\end{figure} 

\subsection{Background Knowledge}
\label{sec-7-2}
The directional and distance based primitives are extracted by
domain-independent mechanisms encoded in the Spatial Visual
System. The agent can query SVS for the value of the primitives for
any object pair in the perceptual field.

\subsection{Acquisition}
\label{sec-7-3}

The acquisition of spatial prepositions is described below as it
relates to the two types of learned knowledge: semantic and
linguistic. The mentor teaches the agent a new spatial relation by
referring to two objects in the world that demonstrate that
relationship. The teaching example \emph{The square is left of the
  circle} is explored in the following sections and depicted in Figure
\ref{fig:leftmap}.

\textbf{Semantic}: After the sentence is parsed and the objects
indexed, SVS is queried for all of the current directional primitive
relations between the specified objects and the values of the distance
primitives. In Figure \ref{fig:leftmap}, three successive teaching
examples are shown; the reference object is the blue circle, and the
three numbered squares are the the primary objects for the three
examples.  The first teaching example exhibits the directional
primitives \texttt{X-less-than}, \texttt{Z-aligned}, and
\texttt{Y-aligned}.  This enables the acquisition of a specific sense
of \emph{left of}, with a strict alignment, with just one
example. However it is possible that the learned information is more
specific than intended. The second example teaches that
\texttt{Y-aligned} or \texttt{Y-less-than} is permissible and the
third example teaches that \texttt{Y-greater-than} is acceptable as
well, forming a disjunction of primitives for the \texttt{Y}
axis. With these additional examples, the agent has a distribution of
distances along the \texttt{X} and \texttt{Y} axes that are used to
calculate a range. This compositional structure, including the
distance ranges and conjunctions and disjunctions of directional
primitives, is stored in semantic memory. The right-side of the figure
displays a representation of this final version of this structure.

Distances from one example are insufficient to learn whether a
distance-based metric is related to the preposition. Given multiple
examples, the distribution of distances is used to determine an
acceptable distance range for the relation. For a relation like
\emph{near} this range may be very restrictive, but for \emph{left
  of}, a large range indicates that distance is not an important
metric for that preposition. The system also records if any examples
are aligned on every axis simultaneously so that the system can
distinguish examples that are explicitly inside or intersecting.

\textbf{Linguistic Mapping}: The preposition linguistic term used to
describe the new relationship is mapped to the learned compositional
structure in semantic memory. In the above example, \emph{left of} is
mapped to this representation and stored into semantic memory, where
it can be accessed when \emph{left of} is used in the future.

\subsection{Spatial Projection}
\label{sec-7-4}

So far we have described how the system acquires and represents
spatial prepositions. This knowledge can be accessed with a lookup
into semantic memory using the linguistic prepositional term whenever
the system needs to know how two objects spatially relate. When
executing an action in the world that attempts to establish the
spatial relation underlying a preposition (such as \emph{put the
  object to the left of the pantry}), the system needs to be able to
\emph{project} that learned preposition to a specific point
(\texttt{X,Y,Z}) in the world.

The current representation provides sufficient information to project
to a point in the world. The legal alignments along each axis are
known as well as the average distance of the examples. When there are
multiple legal alignments for one axis, one is randomly chosen. If the
chosen option is aligned, then a distance of 0 is used and otherwise
the average distance. For example take the situation when projecting
\emph{put the object to the left of the pantry}, using the learned
representation from Figure \ref{fig:leftmap} of \emph{left of}. If the
pantry is located at $(x_i,y_j,z_k)$, the projection point would be
$(x_i-1.7,y_j,z_k)$, representing alignment on the \texttt{Y} and
\texttt{Z} axes and the average displacement on the \texttt{X} axis.

\section{Action Verb Acquisition}
\label{sec-8}
The verb acquisition problem can be decomposed as the acquisition of
knowledge in three distinct but related categories. \emph{Linguistic
  mapping} knowledge allows the agent to associate the linguistic
forms (verbs, prepositions, and noun phrases) in an action command
with an action-concept network. The agent also acquires these
action-concept networks (\emph{semantic} knowledge), which represent a
novel action, its parametrization and goals, and their mutual
constraints. The linguistic and semantic knowledge is represented in
semantic memory. Finally, the agent acquires \emph{procedural
  knowledge} of the novel action, a composition of the primitive
actions that leads to the goal. This knowledge is represented as rules
that encode the action selections necessary to achieve the goal of the
verb. After successful acquisition, the agent can comprehend and
execute verb commands by grounding linguistic forms to objects and
actions.

A typical example of the types of verbs that our system can learn is
\emph{move}, which is a composition of known primitive action verbs
\emph{pick up} and \emph{put down}. On learning the verb \emph{move},
the agent can comprehend the action command \emph{move the red
  triangle to the pantry} by grounding the arguments \emph{the red
  triangle} and \emph{the pantry} to objects present in the
environment and instantiating (and later, executing) the action that
results in \emph{the red triangle in the pantry}. Since, the emphasis
of this work is on acquisition of knowledge through interactive
instruction, we rely on the human instructor to explicitly provide
examples of all knowledge categories described above through natural
language.

\subsection{Characterization}
\label{sec-8-1}
The agent can acquire action verbs that can be characterized along the following dimensions.
\vspace{-0.25cm}
\begin{packeditemize}
\item Parametrization: A single verb word may map to different
  action-concepts based on its argument structure. Some verb commands
  such as \emph{move the red block to the pantry} make the argument
  structure \emph{explicit} by including the object and the location
  in the command. However, in some verb commands such as \emph{store
    the red block}, an argument (\texttt{location} = \texttt{pantry})
  might be \emph{implicit} in the verb \emph{store} itself. The agent
  can learn both categories of verbs.

\item Goal: The learning is currently limited to verbs that have
  perceptible, spatial goal states, such as \emph{the red block is to
    the right of the pantry}.

\item Composition: The agent learns verbs that are compositions of
  known primitive actions, such as \emph{move} which can be executed
  by \emph{picking-up} an object and \emph{putting-down} at a
  position. The agent can learn arbitrary long sequences of
  primitives, if they are useful in achieving the goal.

\end{packeditemize}
\vspace{-0.25cm}
We are interested in \emph{fast} and \emph{general} learning from a
small number of examples of action execution. The agent uses
\emph{chunking}, similar to \emph{explanation-based generalization}
\cite{Rosenbloom1986} to develop casual connections between primitive
actions and the goal of the verb. It extracts rules that are generally
applicable from specific instances of action execution.

\subsection{Background Knowledge}
\label{sec-8-2}
\textbf{Linguistic Mapping}: To ground an action command the agent
must associate the verb and its argument structure to actions and
objects in the environment. This mapping for known primitive verbs is
encoded declaratively in the agent's semantic memory and allows the
agent to access the related action operator which can be instantiated
with objects in the environment. Consider the example shown in Figure
\ref{fig:snet}, Network A. It maps the verb \emph{put} with an
argument structure consisting of a \emph{direct-object} and the object
connected to the verb via the preposition \emph{in} with the operator
\texttt{op\_put-down}. This allows the agent to associate the sentence
\emph{Put a large red block in the table} with an instantiated
operator which will achieve the intended goal.

\textbf{Procedural}: The agent has pre-programmed rules that allow it
to execute the primitive actions in the environment. The actions are
implemented as operators in Soar. An action is defined by its
availability conditions (the preconditions of the action), execution
knowledge (rules that execute action commands in the environment), and
termination conditions (a set of predicates that signify that the goal
of the action has been achieved). The agent maintains a set of all
available primitive actions given the current physical constraints,
object affordances, and the agent's domain knowledge. The agent also
has domain action models (encoded as rules) with which it can simulate
the effect of its actions on the environment during learning.

\subsection{Acquisition}
\label{sec-8-3}
Acquisition of new verbs is integrated with interactive execution of
tasks. If the agent cannot ground an verb command to a known action,
it tries to learn it through agent-initiated situated interactions. We
use the example in Figure \ref{fig:dialog} to demonstrate acquisition
of the verb \emph{store}. 

\begin{figure}[t]
\vskip -0.1in
\begin{center}
\includegraphics[width=0.97\textwidth]{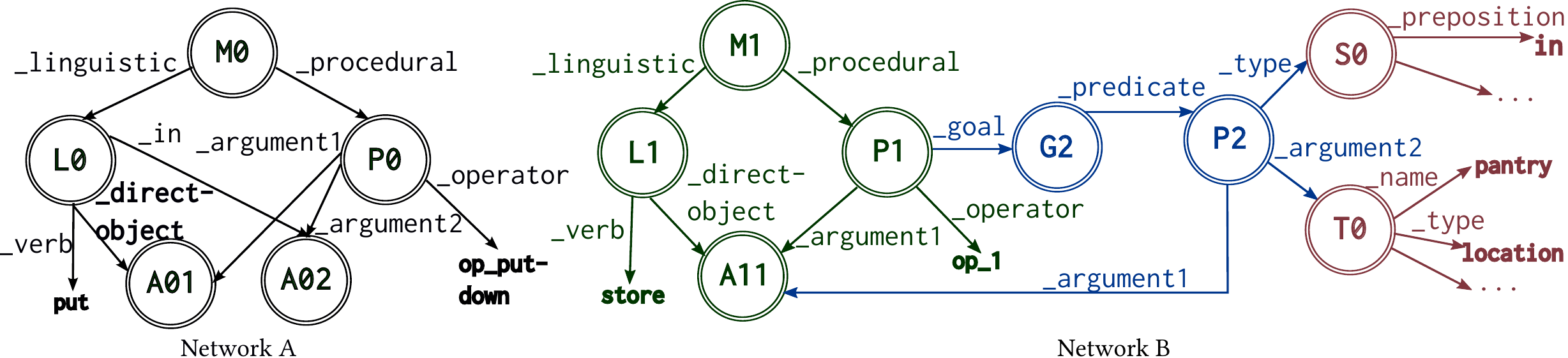}
\vskip -0.05in
\caption{(Left) Pre-encoded linguistic map for
  \emph{put}. (Right) Acquired action-concept network for
  \emph{store}.}
\label{fig:snet}
\end{center}
\vskip -0.25in
\end{figure}

\textbf{Linguistic Mapping}: Using the argument structure of the
action command \emph{Store the orange triangle} (extracted by the
syntactic parser), the agent creates a new mapping in its semantic
memory (shown in Figure \ref{fig:snet}, Network B, nodes (\texttt{M1},
\texttt{L1}, \texttt{A11}, \texttt{P1})). This mapping associates the
novel verb \emph{store} and its \emph{direct-object} to a new operator
(\texttt{op\_1}) and its argument. \texttt{A11} is a slot that can be
filled by an object that satisfies the description (noun-phrase)
connected to the verb as a \emph{direct-object}. corresponds to the
\emph{direct-object} of the verb. The edge \texttt{A11,P1} constrains
the instantiation of \texttt{argument1} to the same object. Future
action commands containing the verb \emph{move} can be indexed to the
action \texttt{op\_1} using this map. After acquiring this mapping,
the agent proposes the operator \texttt{op\_1}. However, since it does
not know how to execute this operator yet, it experiences an
impasse. This impasse causes the agent to begin a new interaction
focusing on learning the verb.

\textbf{Semantic}: The agent prompts the instructor for an explicit
description of the goal (Figure \ref{fig:dialog}, segment
\texttt{G12}). On getting a reply, the agent grounds the utterance to
objects and their spatial relationships. This representation is used
to augment the action-concept network (Figure \ref{fig:snet}, nodes
(\texttt{G2}, \texttt{P2})). The edge \texttt{P2,A11} introduces
constraints on how the goal can be defined on the basis of
instantiation of the verb. This network creates connections between
the linguistic knowledge of the verb and the compositional structure
of the preposition \emph{in} and the \emph{pantry}. These
concepts may have been acquired in previous experiences or may be
learned with the verb acquisition (as in Figure \ref{fig:dialog}).

\textbf{Procedural}: The agent has not yet acquired execution rules
for the verb \emph{store}, hence, the impasse is not resolved. The
agent initiates an interaction with the instructor to acquire a
example execution of the action. Through this interaction,
the instructor decomposes the action into a sequence of primitive
actions (Figure \ref{fig:dialog}, segments \texttt{A13},
\texttt{A14}), which the agent executes in the environment as they are
provided until the goal of the verb is achieved. These interactions
are automatically stored in the agent's episodic memory and are used
for retrospective learning \cite{Mohan2012b}.

Once the goal is achieved, the agent attempts to learn the conditions
under which it should execute the instructed primitive action to
achieve the goal. This process involves using episodic memory to
reconstruct the state that the agent was in when the new verb was
first suggested. From there, the agent continues to use episodic
memory to direct the selection of the instructed actions, and uses its
action-models to simulate the instructed actions. This internal
projection generates an explanation from the initial state to the
goal. From this explanation, the agent extracts the casual
dependencies (via chunking) and learns rules for selecting each one of
the component actions.

\section{Evaluation}
\label{sec-9}
We have described an agent that acquires diverse knowledge including
perceptual classification, spatial composition, and procedural and
semantic knowledge by grounding natural language instructions to the
immediate environment. This establishes our claim that our agent
implemented in Soar supports learning from various sources of
information including the human instructor. The interaction system
provides \emph{flexibility} to support various levels of
instructor/agent control over learning. If the agent is unable to
progress, it poses queries to the instructor and incorporates the
replies in its domain knowledge\footnote[1]{http://youtu.be/\_ktny-h0KXfour}. The
instructor can guide learning by presenting the useful concepts before
the agent is asked to perform a command\footnote[2]{http://youtu.be/9M-rpdXFbgs}.

For evaluating our other claims\footnote[3]{The evaluation was conducted by Soar
  Technology, Ann Arbor}, we created a space of examples for each
category of concepts (noun/adjectives, prepositions, verbs) the agent
can acquire. A random ordering (\emph{trial}) of examples was
generated. The examples from a trial were presented sequentially to
the agent in an \emph{interleaved} training and testing session. The
agent was tested for correctness on every example. In case the example
was unknown to the agent or the agent was incorrect, the example was
added to the training set and relevant instructions were
provided. Trials were repeated until the performance converged (two
successive trials with 100\% performance). We report the results using
the average number of examples required to learn the concepts. The
results were averaged across three runs.

For noun/adjective learning, separate example spaces for properties
color, size, and shape were generated. Each space contained 12 objects
selected randomly from a set that had four distinct colors, two sizes, and
four shapes. The agent was presented with an object and asked for a word
associated with the object for that property. For example, the agent
was asked \emph{What color is this?} for color. The correct word was
provided if the answer was unknown or incorrect. The agent associated
correct colors to 100\% of the objects on observing an average of one
example per color. 100\% performance was achieved after the agent
observed an average of 1.5 examples per size. The agent associated
correct shapes to 96\% of the objects on observing an average of 12.9
examples per shape.

For evaluating preposition learning, six prepositions -- \emph{left},
\emph{right}, \emph{front}, \emph{behind}, \emph{near}, and \emph{far} were
selected. For each preposition, two objects (\emph{red}, \emph{blue})
were arranged in a spatial configuration that was representative for the preposition. The agent was then asked an
evaluation question, \emph{Is the blue object to the right of the red
  object?}. The agent was corrected if it replied incorrectly or
reported that the preposition was unknown. The agent was able to
correctly recognize 93.98\% of 144 spatial arrangement and preposition
associations after an average of 3.17 examples/preposition.

To generate the space for verb acquisition, we created action command
templates from three novel verbs, three for the verb \emph{move} by combining
it with prepositions - \emph{in}, \emph{left of}, and \emph{right of}
and one each for \emph{store} and \emph{discard}. Templates were
instantiated with randomly selected objects (from a set of four) and
locations (a set of four) to generate commands. The initial state of the
arm (\texttt{holding(obj1)} or \texttt{empty}) was also randomly
assigned. If the agent asked for instructions, superfluous actions
were randomly introduced in the instructions. The training converged
after an average of 1.26 examples per action command. We then
generated five random instantiations of every template to test the
agent. The agent executed all instantiations correctly. To understand
the generality in learning, consider the template \emph{move}
\texttt{<obj1>} \emph{to the right of} \texttt{<obj2/location>} for
which there are 64 possible instantiations; with only two examples, the
agent can execute all instantiations.

The data presented above shows that the agent's learning is
\emph{fast}. For various concepts, the agent was able to extract
general knowledge that covered 93.8\% - 100\% of the example space
from very few (1-13) examples. The interactive learning paradigm
allows the agent to be more discriminative while gathering examples,
it gathers few examples for concepts that are easy to learn but
gathers more examples for concepts that are hard to acquire, making
acquisition \emph{efficient}. In noun/adjectives, the agent gathers
more examples for shape than for color or size, as shape is harder to
learn given the perceptual features the agent uses.  For learning
prepositions, the agent adapts interactions to gather more examples
for concepts that are hard such as \emph{near} for which the agent
collects 4.67 examples on an average. This can be compared to the
preposition \emph{behind} (reference object is \emph{directly} behind
the primary object) for which the agent collects only 1
example. Similarly, in verb acquisition, the agent asks for
instructions only when its knowledge is insufficient for making
further progress.

\begin{figure}[t]
\begin{center}
\includegraphics[width=0.47\textwidth]{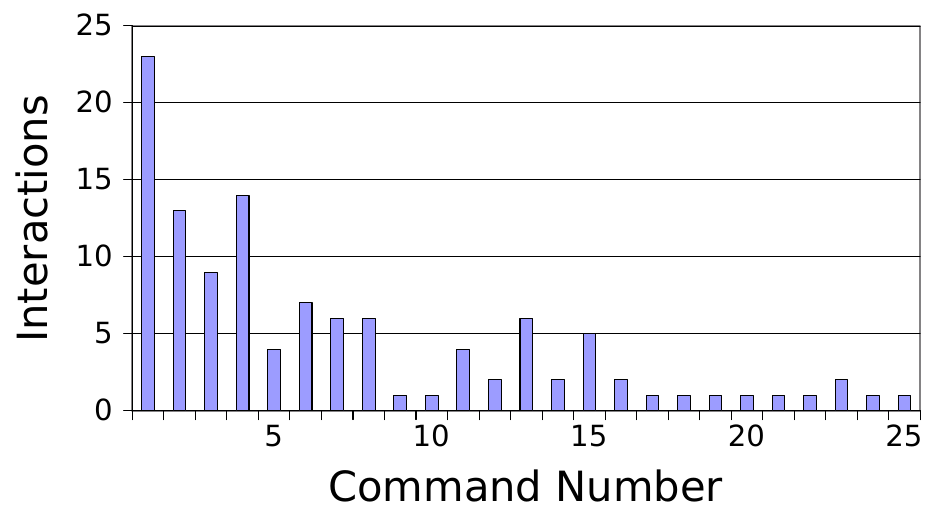}
\includegraphics[width=0.47\textwidth]{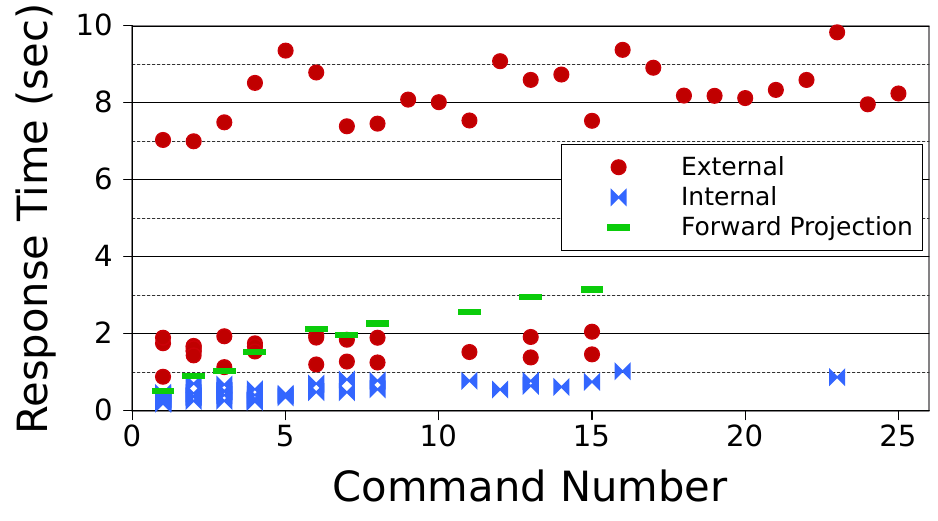}
\vskip -0.1in
\caption{(Left) Number of agent-initiated interactions per action-command. (Right) Agent response time.}
\label{fig:interaction-data}
\end{center}
\vskip -0.3in
\end{figure} 

Figure \ref{fig:interaction-data} (left) shows the agent's performance
while it was learning over a combined space of nine nouns and adjectives,
three prepositions, and three actions. The example space contained
action-commands composed from combining randomly selected verbs,
prepositions, and objects. The agent begins with no prior
knowledge. In the first command, the agent initiates 24 interactions
as it attempts to learn about nouns/adjectives, preposition, and
verbs. As more commands are given, the number of interactions reduce
as the agent generalizes knowledge gathered from previous commands and
applies it in novel situations. Finally, the number of interactions
for a command reduce to one - the utterance from the instructor. This
establishes our claim that agent's learning is \emph{online} and
\emph{incremental}. Figure \ref{fig:interaction-data} (right) plots
the time taken by the agent to generate responses to instructor's
utterance. The responses that involved internal processing were
generated under 1.1 seconds. Responses that involve taking an external
action in the environment and learning from projection take longer,
but can be interrupted and do not impact the reactivity of the
agent. The variation in the external action times are due to the motor
demands of the arm.

\section{Related Work}
\label{sec-10}
Several communities have looked at different challenges involved with
learning grounded representations from knowledge-level human-agent
interaction. The following sections describe the research work from
different communities that is closely related to this paper. 
\subsection{Grounded Language Acquisition}
The problem of language acquisition can be viewed from different
perspectives. Several works have looked at acquisition of grammar
grounded in action plans \cite{Chen2011} and perception
\cite{Matuszek2012}. Our work assumes an English grammar and provides
an approach for grounding linguistic components - words and phrases in perception,
spatial relationships, and action representations. This has been
investigated previously by \emcite{Roy2005}, by \emcite{Gorniak2004}, and
by \emcite{Tellex2011}. Their work has focused on batch learning from
free-form English corpora (and corresponding visual scenes/plans) generated by human
viewers. These mechanisms are robust to user errors and imperfect language use. In contrast, our work aims at
developing learning methods that are online and incremental such that the agent
learns quickly while maintaining an conversation with a human
partner. Interactive learning is useful in acquiring relevant pieces of
knowledge by focusing the conversation on specific components,
resulting in efficient learning. However, our current methods are
susceptible to instruction errors.

\subsection{Human-Robot Dialog}
Research in human-robot interaction has primarily focussed on the
integrative interaction to develop systems that maintain multi-modal
communication with a human partner and solve collaborative
tasks. \emcite{Cantrell2011} demonstrate a natural language understanding
architecture for human-robot interaction that integrates speech recognition,
incremental parsing, incremental semantic analysis and situated
reference resolution. The semantic interpretation of sentences is
based on lambda representations and combinatorial categorial
grammar. They extended the system further to learn definitions from
linguistic instruction. This system can acquire new task knowledge
through task descriptions. In comparison, our agent is a comprehensive learner that
can acquire diverse knowledge including perceptual, spatial, semantic,
procedural, and linguistic knowledge that is grounded in its sensory
perceptions and interactions. 

\subsection{Learning from Human-Agent Interaction}
A substantial work in learning from interaction has looked at
acquisition of control policies from demonstrations or via inverse
reinforcement learning. Learning from knowledge-level interactions
have been addressed by \emcite{Chen2010} who describe a unified agent
architecture for human-robot collaboration that combines natural
language processing and common sense reasoning. They developed a
planning agent that relies on communication with the human to acquire
further information about underspecified tasks. Our agent uses interactive instruction to learn a
wider variety of knowledge, including grounded representations of
language including task execution knowledge.

\emcite{Allen2007a} demonstrate a collaborative task learning agent that
acquires procedural knowledge through a collaborative session of
demonstration, learning, and dialog. The human teacher provides a set
of tutorial instructions accompanied with related demonstrations using
which the agent acquires new procedural knowledge. Although the learning is human demonstration
driven, the agent controls certain aspects of its learning by making
generalizations without requiring the human to provide a large number
of examples. A key distinction of our work is that the initiative of
learning is placed with both the instructor and the agent. The agent
can initiate a learning interaction if its knowledge is insufficient
for further progress, and the instructor can verify the agent's
knowledge by asking relevant questions. This leads to effective and
efficient learning. 

\section{Future Work}
The focus of our future work is on expanding the complexity of tasks
and types of instructions covered by our system. We will also
investigate recovering from instruction error, dealing with perceptual
uncertainty, and learning from instructions situated in historical or
hypothetical context. Some extensions, such as real-time speech
processing and incremental comprehension, would enhance user
interaction, but they would not fundamentally change how the system
learns, so for now, they are not a priority. Our future work will
explore the following dimensions.

\textbf{Complex Tasks}: Along with increasing the number and
types of visual properties we can learn, we are also interested
in physical properties such as the weight of the objects that are
not visually accessible. They require the agent to execute a
sequence of actions to establish their values. We are also
interested in acquiring semantic, hierarchical organization of
objects based on perceptual and functional properties through
instruction. We expect the agent to acquire knowledge such as
\emph{cans are gray cylinders}, or that only some objects can be
picked up. This knowledge could aid in acquiring and generalizing
action affordances and serve as a basis to acquire new proposal
rules. 

The acquisition of prepositions is currently limited to simple binary
spatial relationships between two objects, such as \emph{right of}. We
plan to extend the system to acquire more complex relations involving
multiple objects, such as \emph{between}, to learn contact based
relations, such as \emph{on top}, and to learn logical combinations of
objects and prepositions, such as \emph{clear means all objects
  outside}.

Currently, the agent learns action verbs that are defined by
perceptual features of the final state and are compositions of known
primitive actions. Future work will include verbs that
convey state information (such as \emph{remember}, \emph{belong}) and
verbs which are defined in part by the way in which the underlying
actions are performed (such as \emph{push} versus \emph{move}). We also plan to
expand the types of goal definitions and include the ability to
learn goal definitions automatically through experience.

\textbf{Robustness to Instruction Error}: We currently assume
that the instructor has perfect information about the environment
state and is unlikely to make any instruction errors. However,
this assumption does not hold for instructions in complex tasks
in partially observable environments and for novice
instructors. There are several challenges that have to be
immediately addressed. On the interaction front, we are studying
various interactions that can be useful in corrective
instructions. In corrective instruction, the instructor observes
agent's performance and provides better alternatives
later. Comprehension of such instruction requires the agent to be
able to resolve references to past events. We are also exploring how the
integration of reinforcement learning with instruction and
statistical concept learning can be used for dealing with bad
examples. 

\textbf{Perceptual Uncertainty}: A view of an object from a
single perspective (our current design) occludes several features
that are useful for object identification and
classification. The embodiment of the agent in the
world where it can manipulate objects can be useful in
alleviating this problem. The agent can take several information
gathering actions (such as rotating or moving an object using the
arm) for collecting more information about an object in case there is a
perceptual impasse. Such design poses certain interesting
research questions including what motivates further investigation
as opposed to further interaction with the human. We also intend
to move away from the assumption of complete observability of the
workspace, and investigate how various long-term memories of
experience with the environment can be exploited for providing
the missing sensory information. 

\textbf{Instruction}: We are investigating ways that reduce the number
of interactions required to learn, making learning with instruction
efficient. One such way is to allow the instructor to provide instructions
for situations that slightly deviate from the current state. The
instructor does not have to wait for these situation to arise in the
environment to provide instructions. Interpretation of such
instructions requires the agent to access to sensory information
available from prior experiences with the environment to create
hypothetical situations.

\begin{acknowledgements} 
\noindent
The work described here was supported in part by the Defense Advanced
Research Projects Agency under contract HR0011-11-C-0142. The views
and conclusions contained in this document are those of the authors
and  should not be interpreted as representing the  official policies,
either expressly or implied, of the DARPA or the U.S. Government.

\end{acknowledgements} 

\label{sec-11}
\bibliographystyle{cogsysapa}
\bibliography{bibliography}
\end{document}